\documentclass[unnumsec,webpdf,contemporary,large,namedate]{oup-authoring-template}%

\graphicspath{{Fig/}}

\theoremstyle{thmstyleone}%
\theoremstyle{thmstyletwo}%
\theoremstyle{thmstylethree}%

\usepackage{booktabs}
\usepackage{placeins} 
\usepackage{tabularx}

\journaltitle{Article}  
\copyrightyear{2026}
\pubyear{2026}
\access{Advance Access Publication Date: Day Month Year}
\appnotes{Paper}

\begin{document}


\firstpage{1}


\title[Article Title]{LaCoGSEA: Unsupervised deep learning for pathway analysis via latent correlation
}

\author[1]{Zhiwei Zheng}
\author[1,$\ast$]{Kevin Bryson\ORCID{0000-0002-1163-6368}}

\authormark{Zheng et al.}

\address[1]{\orgdiv{School of Computing Science}, 
             \orgname{University of Glasgow}, 
             \postcode{G12 8RZ}, 
             \state{Glasgow}, 
             \country{United Kingdom}}

\corresp[$\ast$]{Corresponding author: 
\href{mailto:Kevin.Bryson@glasgow.ac.uk}{Kevin.Bryson@glasgow.ac.uk}}




\abstract{\textbf{Motivation:} Pathway enrichment analysis is widely used to interpret gene expression data. Standard approaches, such as GSEA, rely on predefined phenotypic labels and pairwise comparisons, which limits their applicability in unsupervised settings. Existing unsupervised extensions, including single-sample methods, provide pathway-level summaries but primarily capture linear relationships and do not explicitly model gene–pathway associations. More recently, deep learning models have been explored to capture non-linear transcriptomic structure. However, their interpretation has typically relied on generic explainable AI (XAI) techniques designed for feature-level attribution. As these methods are not designed for pathway-level interpretation in unsupervised transcriptomic analyses, their effectiveness in this setting remains limited.
\newline
\textbf{Results:} To bridge this gap, we introduce LaCoGSEA (Latent Correlation GSEA), an unsupervised framework that integrates deep representation learning with robust pathway statistics. LaCoGSEA employs an autoencoder to capture non-linear manifolds and proposes a global gene--latent correlation metric as a proxy for differential expression, generating dense gene rankings without prior labels. We demonstrate that LaCoGSEA offers three key advantages: (i) it achieves improved clustering performance in distinguishing cancer subtypes compared to existing unsupervised baselines; (ii) it recovers a broader range of biologically meaningful pathways at higher ranks compared with linear dimensionality reduction and gradient-based XAI methods; and (iii) it maintains high robustness and consistency across varying experimental protocols and dataset sizes. Overall, LaCoGSEA provides state-of-the-art performance in unsupervised pathway enrichment analysis.
\newline
\textbf{Availability and implementation:} https://github.com/willyzzz/LaCoGSEA.}

\maketitle
\section{Introduction}

Pathway enrichment analysis has become a central paradigm \citep{PEA1, PEA2} for interpreting high-dimensional transcriptomic data, as it aggregates dispersed gene-level signals into coherent biological processes. Among existing approaches, Gene Set Enrichment Analysis (GSEA) \citep{PEA2, OriginalGSEA} is widely regarded as the standard for uncovering coordinated transcriptional programs. However, the applicability of GSEA is fundamentally constrained by its dependence on supervised paradigms \citep{GSEA_limitation1, GSEA_limitation2, GSEA_limitation3}. The algorithm strictly requires a priori clinical labels to compute the differential expression metrics necessary for generating a pre-ranked gene list. Consequently, in exploratory analyses of large and heterogeneous transcriptomic datasets, sample labels are often abundant but poorly aligned with the research question, forcing exhaustive and inefficient pairwise contrasts that limit unsupervised discovery. As a result, the classical GSEA framework is poorly suited for unsupervised discovery-driven analyses, revealing a significant methodological gap.

In such cases, two broad classes of approaches have been commonly used to explore latent biological structure in transcriptomic data. First, unsupervised or single-sample pathway scoring methods, such as GSVA \citep{GSVA} and ssGSEA \citep{ssgsea}, aim to directly estimate pathway activities from individual expression profiles. While conceptually appealing, these methods rely on intra-sample gene rankings and can be sensitive to technical noise, sequencing depth, and expression dropouts \citep{ssmethods_limit1, ssmethods_limit2}, particularly in heterogeneous datasets. Second, dimensionality reduction techniques, including PCA \citep{PCA, PCA2} and NMF \citep{NMF}, are frequently applied to learn low-dimensional representations of gene expression data, followed by post hoc pathway enrichment to interpret the resulting components. However, linear methods enforce strong constraints such as orthogonality or statistical independence \citep{Linear_limit1, Linear_limit2}. Since biological pathways often overlap, imposing such constraints can fragment coherent biological programs into separate components, limiting their ability to capture the complex, non-linear dependencies inherent in transcriptomic landscapes.

Unsupervised deep learning, offers a compelling alternative by learning non-linear data manifolds. However, the black-box nature of the latent space limits the interpretability
of learned representations. To interpret these latent features, recent studies have applied gradient-based explainable AI (XAI) methods \citep{XAI1, XAI2, XAI3, XAI4}. Here, we identify a critical conflict: these XAI methods are designed for prediction tasks and mathematically optimize for sparsity, identifying the minimum subset of features sufficient for prediction. Biological pathways, in contrast, are distinctively redundant and dense, relying on the coordinated expression of large gene networks. Applying sparse XAI attribution to pathway-level signal aggregation can impair rank-based enrichment, as GSEA relies on coordinated high-ranking signals across many genes within a pathway rather than a few isolated features.

\begin{figure*}
    \centering
    \includegraphics[width=\linewidth]{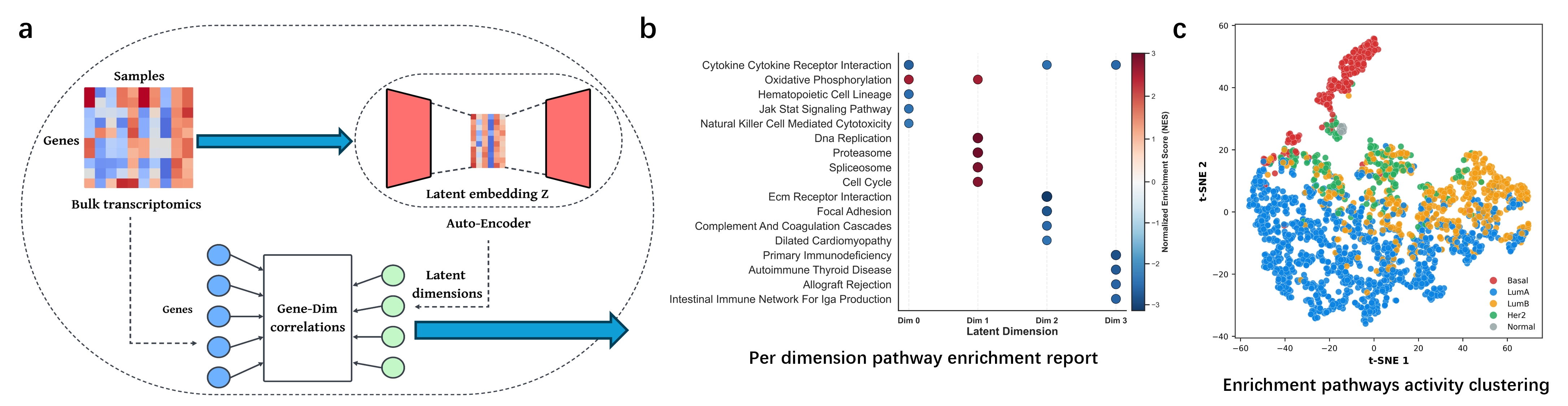}
    \caption{
    Schematic overview of the LaCoGSEA framework. (a) Bulk transcriptomic data are first compressed into low-dimensional latent embeddings using a deep autoencoder. Gene–dimension correlations are then computed between the original gene expression and each latent dimension, producing a gene–dimension correlation map. These correlations are used to generate pre-ranked gene lists for standard Gene Set Enrichment Analysis (GSEA). (b) Application to the SCAN-B dataset with latent dimension $D=4$. For each latent dimension, the top 5 enriched KEGG pathways are reported, which shows with different biology mechanisms. (c) The overall sample-level pathway activity matrix for SCAN-B ($D=4$) is generated and visualized using t-SNE, illustrating coherent separation of samples based on pathway programs.
    }
    \label{fig:LaCoGSEA}
\end{figure*}

To address these challenges, we introduce LaCoGSEA (Fig.~\ref{fig:LaCoGSEA}), a novel framework that enables unsupervised pathway enrichment through an explicit pre-ranking mechanism. LaCoGSEA leverages a deep autoencoder to capture non-linear structure in transcriptomic data, while avoiding complex gradient-based interpretation. Instead, we quantify global correlations between latent dimensions and gene expression as a proxy for biological activity, yielding dense and robust gene rankings that preserve correlated network structure. These rankings allow the direct application of the statistically rigorous GSEA algorithm in a fully unsupervised setting, effectively combining the representational power of deep learning with the interpretability of classical pathway analysis. Across diverse biological benchmarks, LaCoGSEA demonstrates improved pathway recovery, subtype stratification, and robustness under limited sample sizes. Notably, simple global correlation metrics outperform gradient-based explanation methods like SHAP \citep{SHAP} and DeepLIFT \citep{Deeplift}, in retrieving biologically meaningful pathways, highlighting the importance of capturing dense, co-regulated programs rather than sparse driver features.

\section{Materials and methods}\label{sec2}

\subsection{Autoencoder-based latent representation}

We consider bulk transcriptomic data represented as a gene expression matrix $X \in \mathbb{R}^{G \times N}$, where $G$ denotes the number of genes and $N$ the number of samples. To capture non-linear biological dependencies, we treat each sample $x_i \in \mathbb{R}^G$ (columns of $X$) as an input vector and employ a deep autoencoder framework. The framework consists of an encoder $f_\phi: \mathbb{R}^G \to \mathbb{R}^D$ ($D \ll G$) that maps high-dimensional profiles to a compressed latent manifold, and a decoder $g_\psi: \mathbb{R}^D \to \mathbb{R}^G$ that reconstructs the input. The network parameters $\theta = \{\phi, \psi\}$ are jointly optimized to minimize the reconstruction error. To preserve correlated gene modules, we incorporate an Elastic Net regularization term (combining $L_1$ and $L_2$ penalties) into the objective function:
\begin{equation}
\mathcal{L}(\theta) = \frac{1}{N} \sum_{i=1}^{N} \| x_i - g_\psi(f_\phi(x_i)) \|_2^2 + \lambda_1 \|\theta\|_1 + \lambda_2 \|\theta\|_2^2,
\label{eq:ae_loss}
\end{equation}
where $z_i = f_\phi(x_i)$ represents the intrinsic biological state, and $\lambda_1, \lambda_2$ control the sparsity and smoothness constraints, respectively. This hybrid regularization strategy effectively mitigates overfitting.

\subsection{Gene--dimension correlation and pre-ranking}

To interpret the latent features learned by the autoencoder, we propose a global attribution strategy. Unlike linear models with explicit loading matrices, non-linear AEs require a post-hoc association metric to map latent states back to gene space. We utilize the Pearson correlation coefficient $\rho_{j,k}$ between the expression vector of gene $j$ and the activation vector of latent dimension $k$ across the cohort. This results in a global correlation matrix $\mathbf{P} \in \mathbb{R}^{G \times D}$.

Crucially, this metric captures intrinsic co-regulation patterns without requiring phenotypic labels or control groups, effectively serving as an unsupervised proxy for differential expression. To prepare for downstream enrichment analysis, we generate a pre-ranked gene list $L_k$ for each dimension $k$ by sorting genes based on their correlation coefficients:
\begin{equation}
L_k = \{ g_{(1)}, g_{(2)}, \dots, g_{(G)} \} \quad \text{s.t.} \quad \rho_{(1),k} \ge \rho_{(2),k} \ge \dots \ge \rho_{(G),k}.
\label{eq:ranking}
\end{equation}
By retaining the sign of $\rho_{j,k}$, this ranking preserves the directionality of biological regulation, where positive and negative values correspond to up- and down-regulation relative to the latent process. These $D$ independent ranked lists $\{L_1, \dots, L_D\}$ serve as the input for standard GSEA.

\subsection{Downstream pathway analysis}

The output of the gene--dimension correlation analysis is a set of $D$ ordered gene lists, one for each latent dimension. We perform downstream analysis in two stages: functional annotation via enrichment analysis and sample-level pathway characterization.

\subsubsection{Functional annotation via enrichment analysis}

To decipher the biological meaning of each latent dimension and evaluate the model's ability to capture target biological processes, we apply standard Gene Set Enrichment Analysis (GSEA) to the pre-ranked gene lists of all latent dimensions. For a given dimension $k$, the ordered gene list is tested against reference gene set databases (e.g., KEGG, GO). A pathway $P$ is considered successfully ``detected'' by dimension $k$ if it achieves statistical significance (FDR $< 0.05$). Since the autoencoder may capture a specific biological signal across multiple latent dimensions (or primarily in a single specific dimension), we evaluate the global performance of the model for a target pathway $P$ by identifying its best ranking across the latent space. Specifically, we define the model-level rank of pathway $P$ as the minimum rank observed across all dimensions where the pathway is significant:
\begin{equation}
\text{Rank}_{model}(P) = \min_{k \in \{1,\dots,D\} \mid \text{FDR}_{P,k} < 0.05} (\text{Rank}_{P,k}).
\end{equation}
If a pathway is not significantly enriched in any dimension, it is assigned a penalty rank to reflect detection failure. This metric, utilized in our benchmarking, assesses how effectively the model prioritizes relevant biological signals compared to baseline methods.

\subsubsection{Sample-level pathway characterization}

A single biological pathway is often regulated by complex mechanisms distributed across multiple latent dimensions. To synthesize these dispersed signals, we aggregate the latent representations into a holistic pathway score. First, we construct a dense weighting matrix $W \in \mathbb{R}^{D \times M}$, where each entry $W_{k,p}$ corresponds to the Normalized Enrichment Score (NES) of pathway $p$ in latent dimension $k$. We retain the full NES matrix to preserve the continuous spectrum of biological associations. The activity score $A_{i,p}$ for pathway $p$ in sample $i$ is defined as the dot product between latent representation and pathway-specific weight vector:
\begin{equation}
A_{i,p} = \sum_{k=1}^{D} z_{i,k} \cdot W_{k,p},
\label{eq:summation}
\end{equation}
where $z_{i,k}$ is the activation of the $k$-th latent dimension. This formulation ensures that latent dimensions with stronger biological associations contribute proportionally more to the final score.

\subsection{Experimental setup}\label{subsec1}

\subsubsection{Datasets and preprocessing}

To rigorously evaluate the performance and generalizability of LaCoGSEA, we collected a comprehensive set of transcriptomic datasets spanning diverse biological contexts, tissue types, and sequencing platforms (summarized in Table~\ref{tab:datasets}). We utilized three large-scale cancer cohorts (SCAN-B \citep{SCANB_PAPER1, SCANB_PAPER2}, METABRIC \citep{METABRIC_PAPER}, and TCGA-Lung \citep{TCGA}) as the primary datasets for model characterization and biological validation. Additionally, five independent datasets \citep{GSE1, GSE2, GSE3, GSE4, GSE5} from the Gene Expression Omnibus (GEO) were curated to benchmark the pathway ranking metrics across various conditions, including neurodegenerative diseases, immune responses, and metabolic disorders. Regarding data preprocessing, we applied a consistent strategy across all datasets to ensure comparability. Raw gene expression values like counts or intensities were transformed using $\log_2(x+1)$ to stabilize variance and reduce the impact of extreme outliers.

\begin{table}[t]
\centering
\caption{Summary of transcriptomic datasets used in this study.}
\label{tab:datasets}

\small
\setlength{\tabcolsep}{6pt}
\renewcommand{\arraystretch}{1.25}

\begin{tabularx}{\columnwidth}{@{}X l r r@{}}
\toprule
Dataset (Condition) & Data Format & Samples ($N$) & Genes ($G$) \\
\midrule
SCAN-B(Breast Cancer)        & RNA-seq     & 7,429 & 16,736 \\
METABRIC(Breast Cancer)     & Microarray  & 1,992 & 16,736 \\
TCGA Lung(NSCLC)             & RNA-seq     & 1,128 & 20,531 \\
GSE10846(DLBCL)             & Microarray  & 420   & 22,880 \\
GSE48350(Alzheimer's)      & Microarray  & 253   & 22,880 \\
GSE11375(Trauma)            & Microarray  & 184   & 22,880 \\
GSE126848(Liver)         & RNA-seq     & 30    & 15,911 \\
GSE116250(Heart)            & RNA-seq     & 51    & 26,243 \\
\bottomrule
\end{tabularx}

\vspace{3pt}
\footnotesize
\textit{NSCLC}: Non-Small Cell Lung Cancer; 
\textit{DLBCL}: Diffuse Large B-Cell Lymphoma.

\end{table}

\subsubsection{Baseline methods for comparison}

To benchmark LaCoGSEA, we compared it with a set of representative baseline methods spanning linear dimensionality reduction, deep learning attribution, and unsupervised pathway activity inference. Principal Component Analysis (PCA) was included as a linear baseline to enable a direct comparison with autoencoder-based representations by controlling the latent dimensionality to be the same. Gene rankings were derived using absolute loading coefficients (PCA\_Weights) and a correlation-based strategy (PCA\_Corr), allowing a controlled comparison between weight-based and correlation-based attribution. To contrast the proposed global correlation metric with established explainable AI approaches, we evaluated two gradient-based methods, SHAP and DeepLIFT, with sample-level attributions aggregated into global gene rankings. As a supervised reference, we included Standard\_DE (standard t-test), which provides a reference on performance when phenotype labels are available. For pathway-level activity inference and downstream clustering, we compared our inferred pathway activities with widely used unsupervised single-sample methods, GSVA and ssGSEA, using standard implementations from the R package GSVA.

\section{Results}\label{sec3}

\subsection{Pathway detection under saturation}

To evaluate the capacity of latent representations to capture biological information, we performed a saturation analysis on the SCAN-B and METABRIC cohorts. We compared the deep autoencoder (AE) against Principal Component Analysis (PCA) by measuring the number of statistically significant pathways identified across increasing latent dimensions ($D \in {1, 2, 4, 8, \dots, 128}$). Crucially, to control for the increasing multiple hypothesis testing burden, we applied a strict Bonferroni correction (threshold $p < 0.05/D$).

\begin{figure*}
    \centering
    \includegraphics[width=\linewidth]{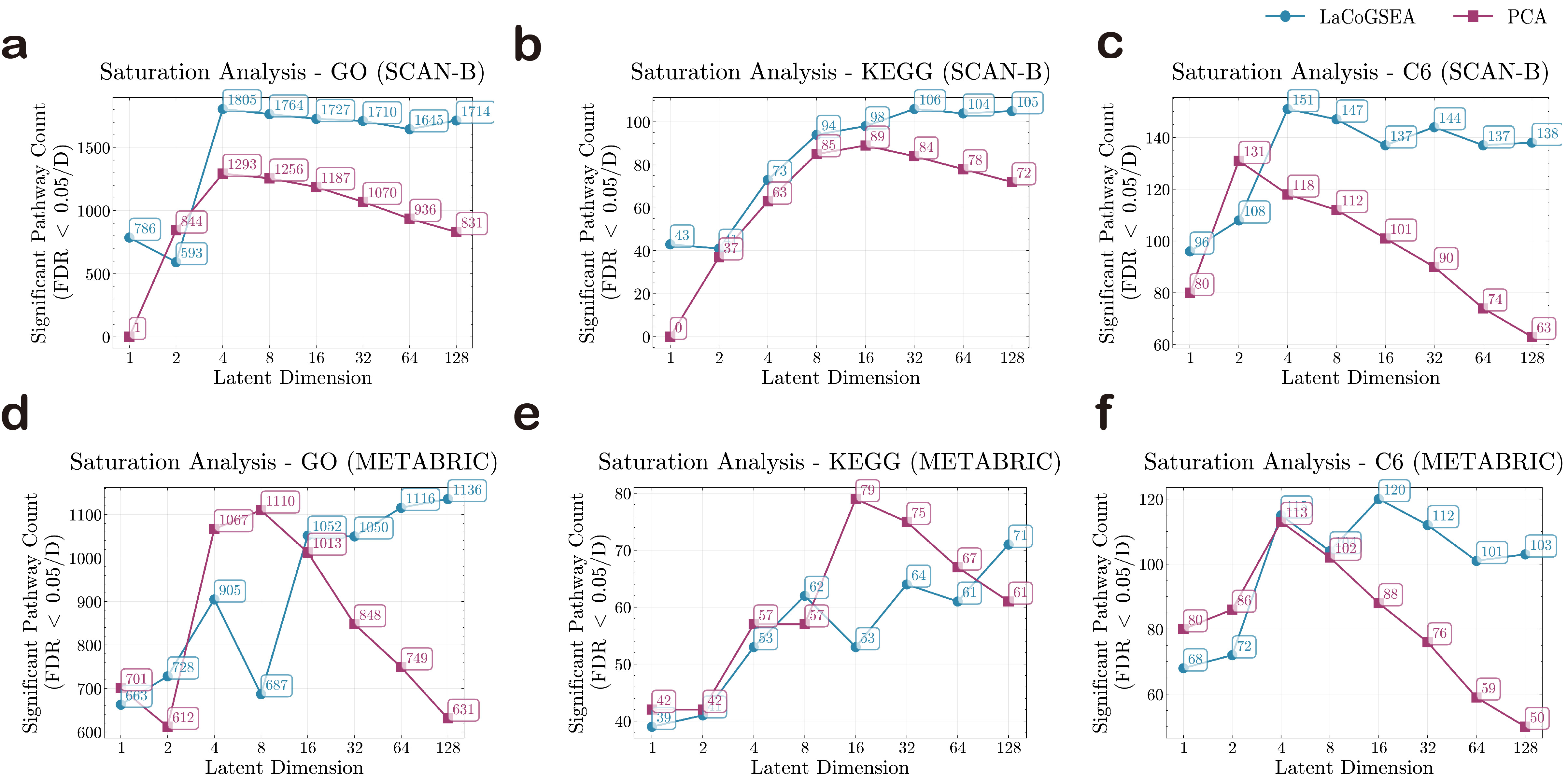}
    \caption{Saturation analysis reveals superior pathway detection capacity of the LaCoGSEA compared to PCA.
(a–f) Comparison of the number of significant pathways detected by AE (blue) and PCA (purple) across increasing latent space dimensions ($D \in {1, 2, 4, 8, \dots, 128}$). Analysis was performed on SCAN-B (a, b, c) and METABRIC (d, e, f) cohorts using three gene set collections: GO Biological Processes, KEGG, and C6 Oncogenic Signatures. Significance was defined using a strict Bonferroni-corrected threshold (FDR $< 0.05/D$).}
    \label{fig:Saturation analysis}
\end{figure*}

In the large-scale SCAN-B dataset, the AE demonstrated superior stability and discovery capacity across all gene set collections (GO, KEGG, and C6; Fig.~\ref{fig:Saturation analysis}a--c). Once the latent dimension exceeded a minimal threshold ($D \ge 4$), the AE consistently identified a high volume of unique pathways and reached a stable saturation plateau. This advantage was particularly evident in the C6 oncogenic signatures (Fig.~\ref{fig:Saturation analysis}c), where the AE maintained robust detection (stabilizing at 138 signatures) while PCA performance degraded notably at higher dimensions (dropping to $<70$ at $D=128$). This performance drop aligns with the known limitations of linear dimensionality reduction in high-dimensional biological data. Unlike the AE, PCA enforces strict orthogonality, a constraint that may limit the effective separation of overlapping biological processes, such as oncogenic cascades into distinct latent components. Consequently, linear components often fail to retain sufficient signal strength to surpass the rigorous Bonferroni threshold at higher dimensions.

In the METABRIC cohort (Fig.~\ref{fig:Saturation analysis}d--f), while the sample size limited the magnitude of improvement, the AE still exhibited a more robust trajectory with less fluctuation than PCA. To verify that the AE's high discovery rate was not driven by overfitting to noise, we performed a negative control analysis using synthetic Gaussian noise (Supplementary Fig. S1). Both methods yielded exactly zero significant pathways, indicating that the increased sensitivity of the AE reflects genuine biological signal recovery rather than false-positive inflation.

Given the consistency and interpretability of KEGG across datasets, we focus subsequent analyses on KEGG pathways. Across datasets and pathway collections, the number of significant KEGG pathways increased rapidly at low dimensionalities and stabilized over a broad range of higher dimensions. We therefore selected $D=64$ as a representative operating point within this saturation regime for all downstream analyses.

\begin{figure*}
    \centering
    \includegraphics[width=\linewidth]{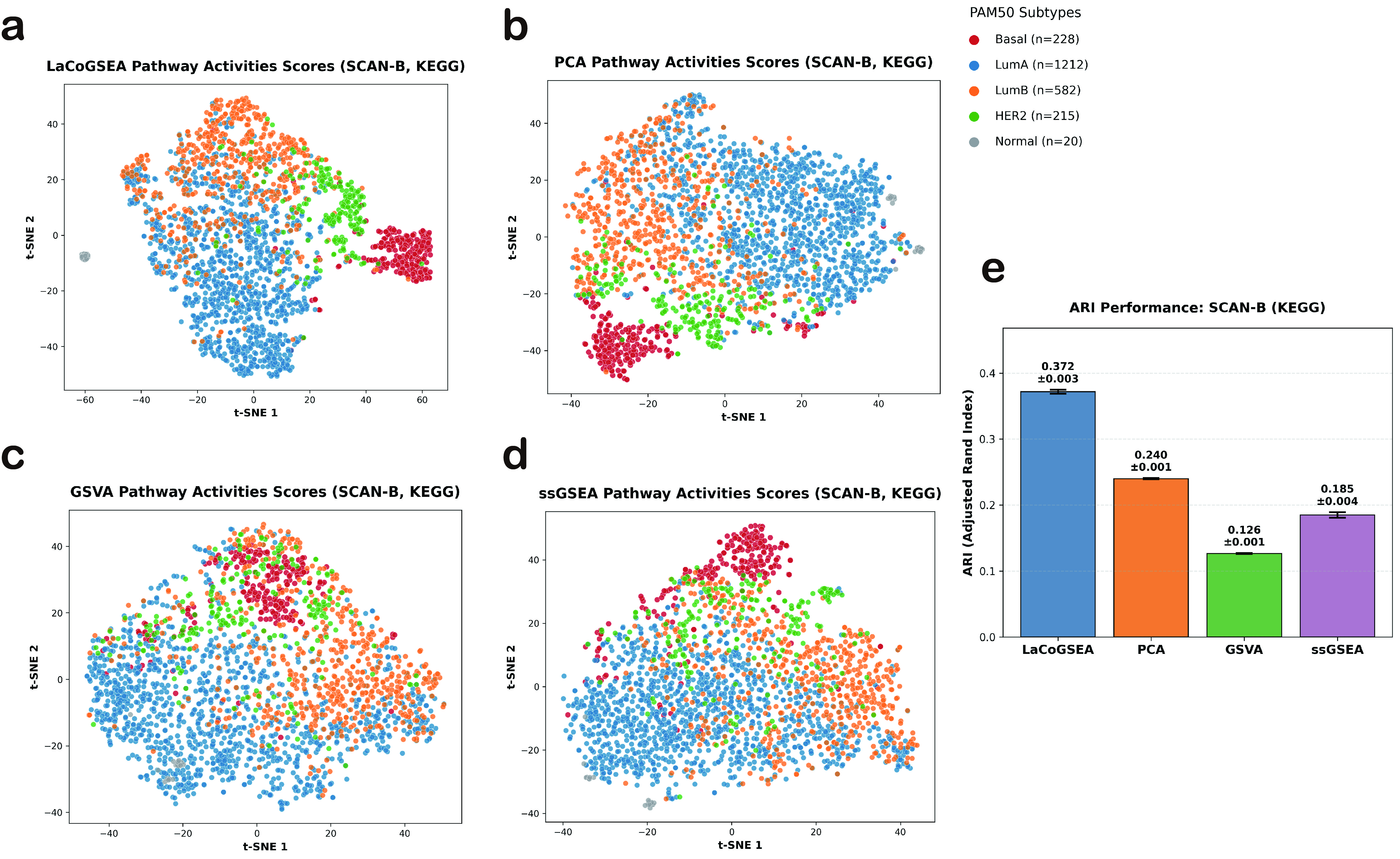}
    \caption{Latent pathway activity inference enhances breast cancer subtype clustering.  (a–d) t-SNE visualizations of pathway activity scores generated by (a) LaCoGSEA, (b) PCA, (c) GSVA, and (d) ssGSEA. Points are colored by PAM50 subtypes (Basal, LumA, LumB, HER2, Normal). (e) Bar chart showing the Adjusted Rand Index (ARI) for clustering performance on the SCAN-B cohort using KEGG pathways. The LaCoGSEA pathway activity matrix achieves the highest concordance with PAM50 clinical subtypes (ARI = 0.372), significantly outperforming PCA (0.240), GSVA (0.126), and ssGSEA (0.185). Error bars represent standard deviation across bootstrap iterations.}
    \label{fig:SCAN-B pathway activity}
\end{figure*}

\begin{figure*}[t]
    \centering
    \includegraphics[width=\textwidth]{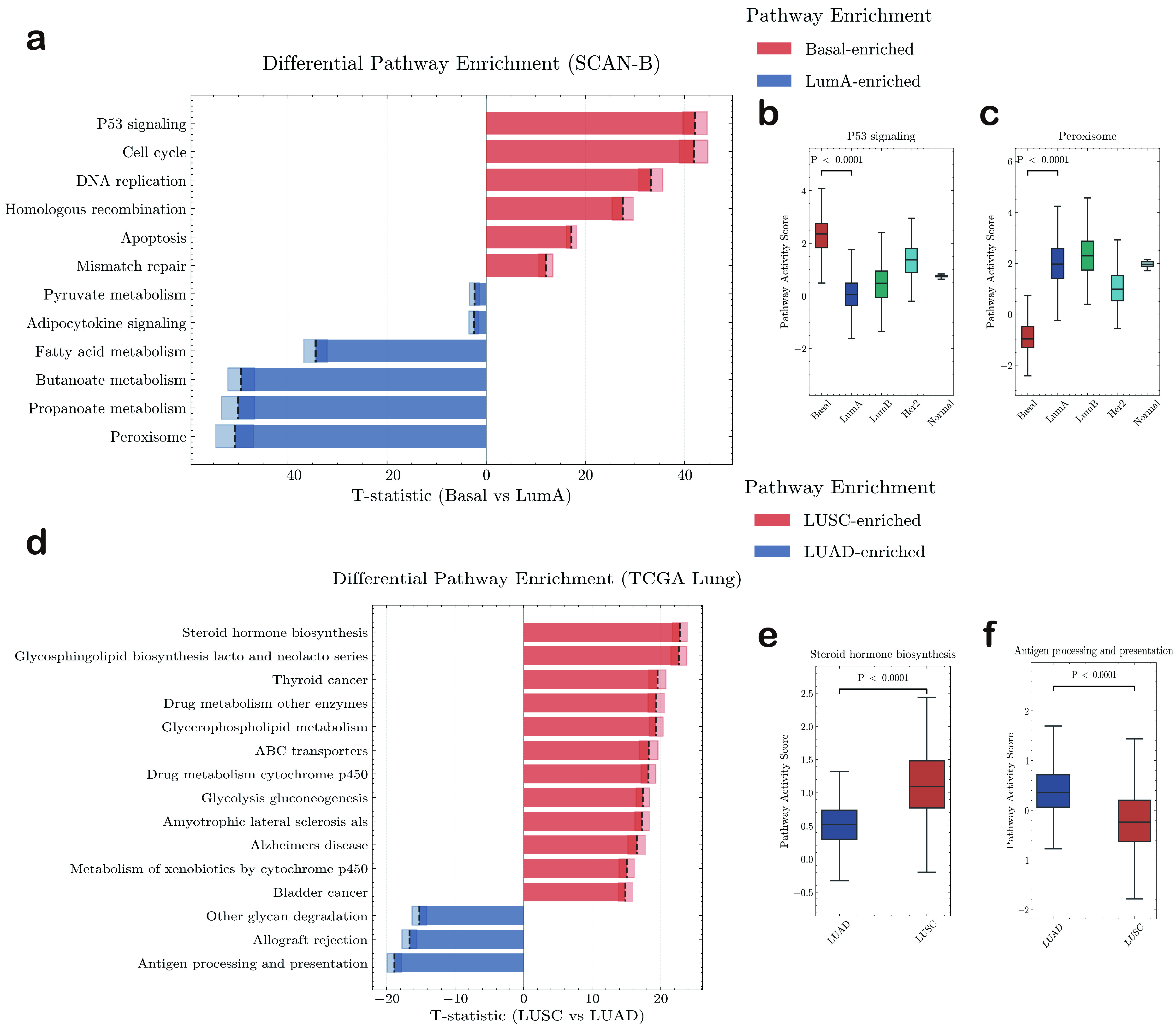} 
    \caption{\textbf{LaCoGSEA identifies distinct metabolic and proliferative drivers across breast and lung cancer subtypes.}
    \textbf{(a)} Differential pathway enrichment analysis between Basal-like and Luminal A subtypes in the SCAN-B cohort. The model captures a fundamental metabolic switch: Basal tumors (red) are dominated by proliferation and DNA repair mechanisms (e.g., \textit{P53 signaling}, \textit{Cell cycle}), whereas LumA tumors (blue) are characterized by oxidative and lipid metabolism (e.g., \textit{Peroxisome}, \textit{Fatty acid metabolism}).
    \textbf{(b--c)} Box plots of inferred pathway activity scores validating the enrichment results. \textbf{(b)} \textit{P53 signaling} is significantly elevated in Basal samples ($P < 0.0001$), while \textbf{(c)} \textit{Peroxisome} activity is significantly higher in Luminal subtypes, consistent with their dependency on fatty acid oxidation.
    \textbf{(d)} Differential pathway signatures distinguishing Lung Squamous Cell Carcinoma (LUSC) from Lung Adenocarcinoma (LUAD) in the TCGA dataset. LaCoGSEA identifies \textit{Steroid hormone biosynthesis} as a primary driver for LUSC.
    \textbf{(e--f)} Validation of subtype-specific activity scores. \textbf{(e)} \textit{Steroid hormone biosynthesis} shows robust upregulation in LUSC, while \textbf{(f)} \textit{Antigen processing and presentation} is significantly enriched in LUAD ($P < 0.0001$, Wilcoxon rank-sum test). Error bars represent the interquartile range.}
    \label{fig:validation_combined}
\end{figure*}

\begin{figure*}
    \centering
    \includegraphics[width=0.9\linewidth]{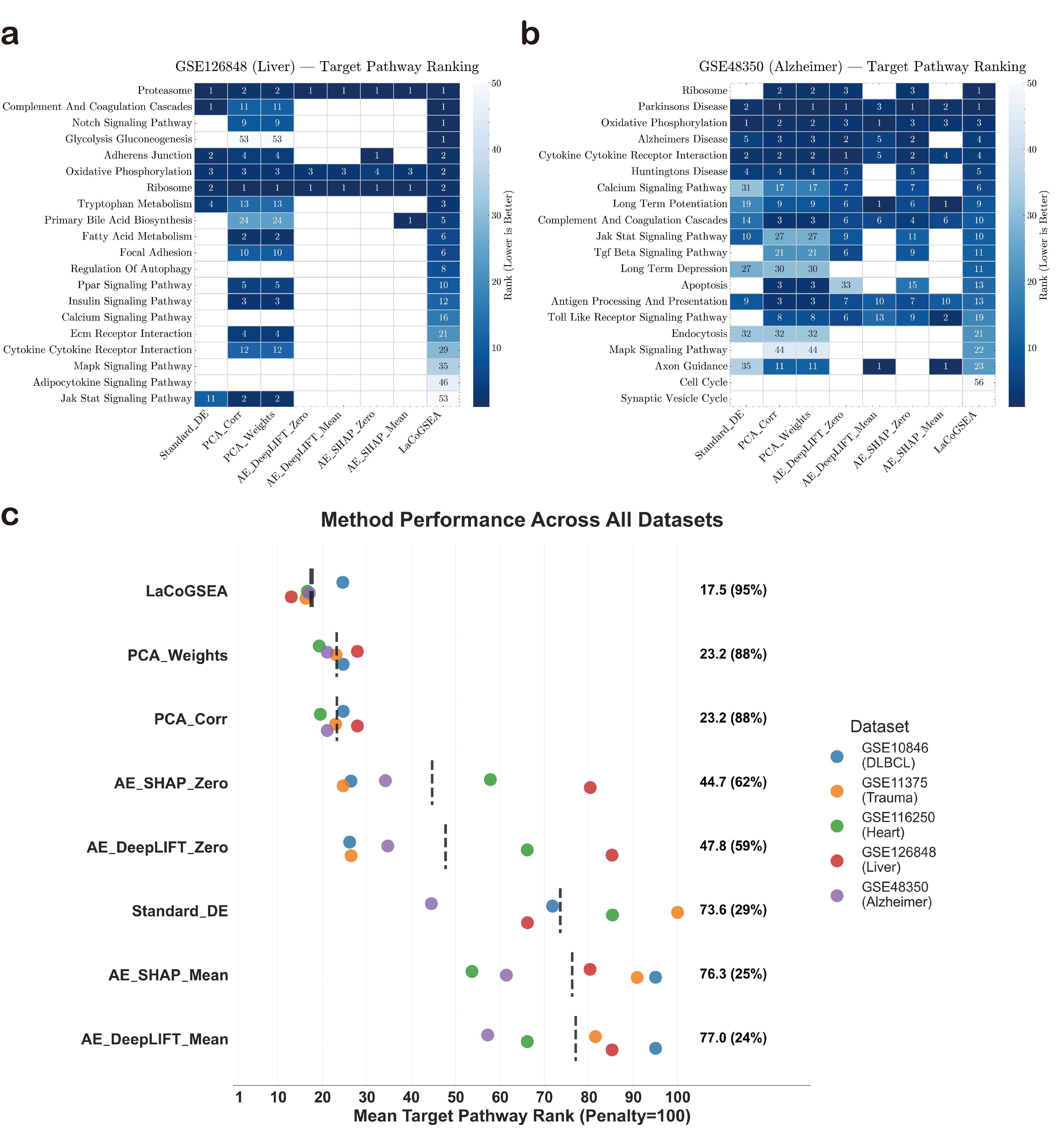}
    \caption{
    LaCoGSEA with global correlation metrics outperforms linear and gradient-based methods in target pathway prioritization.
    (a--b) Heatmaps displaying the ranking of target pathways across different computational methods for
    (a) GSE126848 (Liver Disease) and (b) GSE48350 (Alzheimer's Disease).
    Rows represent known target pathways, and columns represent methods.
    The color intensity represents the rank (darker blue indicates a lower/better rank).
    LaCoGSEA consistently assigns the lowest ranks to validated mechanisms compared to
    PCA baselines (PCA\_Corr, PCA\_Weights) and deep learning attribution methods
    (AE\_DeepLIFT, AE\_SHAP).
    (c) Summary of method performance across five independent benchmarking datasets.
    The dot plot shows the Mean Target Pathway Rank (lower is better) for each method.
    If a target pathway was not detected, a penalty rank of 100 was assigned.
    The vertical black line indicates the mean rank across all datasets.
    LaCoGSEA achieves the best overall performance (Mean Rank $\approx 17.5$),
    significantly outperforming standard DE and interpretability-based AE variants.
    }
    \label{fig:cross_performance_five_datasets}
\end{figure*}

\subsection{Latent pathway activities enable robust subtype stratification and mechanistic discovery}

While the saturation analysis confirms that the LaCoGSEA retrieves a higher volume of pathway signals, a critical question remains: do these additional signals represent meaningful biological heterogeneity? To address this, we transformed the latent representations ($D=64$) into inferred pathway activity profiles by combining latent embeddings with their corresponding dimension-wise enrichment scores, thereby summarizing pathway-associated signals across all latent dimensions for each sample. We then evaluated the biological relevance of these inferred activities through unsupervised clustering and differential analysis on the SCAN-B (Breast cancer) and TCGA-NSCLC (Lung cancer) cohorts.

\subsubsection{Unsupervised stratification performance}
We first assessed whether the inferred pathway activities could recover known molecular subtypes without supervision. Using K-means clustering on the standardized pathway activity matrix, we compared the concordance between the identified clusters and the PAM50 clinical labels in the SCAN-B cohort. LaCoGSEA achieved the highest clustering performance, with an Adjusted Rand Index (ARI) of 0.372, outperforming linear PCA (0.240) as well as standard single-sample enrichment methods including GSVA (0.126) and ssGSEA (0.185) (Fig.~\ref{fig:SCAN-B pathway activity}e).

Visually, the t-SNE projection of the pathway activities revealed a clear separation of Basal-like, Luminal A, and HER2 subtypes (Fig.~\ref{fig:SCAN-B pathway activity}a). In contrast, embeddings generated by GSVA and PCA showed considerable overlap across subtypes (Fig.~\ref{fig:SCAN-B pathway activity}b-d). This distinction underscores a key advantage of our framework: unlike traditional enrichment paradigms that typically rely on binary contrasts with experiment and control groups or sample-independent scoring, the AE's non-linear manifold inherently models the global structure of the data. This allows the model to capture multi-directional heterogeneity across numerous subtypes simultaneously, regardless of the number of classes, thereby resolving complex inter-tumor patterns that remain conflated in linear or pairwise approaches.

\subsubsection{Decoupling metabolic and proliferative drivers in breast and lung cancers}
To understand the specific biological mechanisms driving this superior stratification, we performed a differential pathway analysis between the most distinct subtypes in both cohorts (Figure~\ref{fig:validation_combined}).

In the SCAN-B breast cancer cohort (Fig.~\ref{fig:validation_combined}a-c), comparing Basal-like against Luminal A subtypes revealed a stark dichotomy between proliferation and metabolism. The model correctly identified \textit{P53 signaling}, \textit{Cell cycle}, and \textit{DNA replication} as the top drivers of the aggressive Basal subtype. Conversely, Luminal A tumors were characterized by the upregulation of \textit{Peroxisome}, \textit{Propanoate metabolism}, and \textit{Fatty acid metabolism}. This aligns with the established metabolic switch in breast cancer, where aggressive basal tumors rely on glycolysis and rapid replication, while well-differentiated luminal tumors maintain dependency on lipid oxidation.

We further extended this validation to the TCGA Lung Cancer cohort to distinguish Lung Squamous Cell Carcinoma (LUSC) from Lung Adenocarcinoma (LUAD) (Fig.~\ref{fig:validation_combined}d). LaCoGSEA identified \textit{Steroid hormone biosynthesis} and \textit{Glycosphingolipid biosynthesis - lacto and neolacto series} as the top signatures enriched in LUSC. This enrichment generally aligns with the squamous differentiation lineage, which typically involves complex remodeling of cell membrane lipids and glycoconjugates. In contrast, LUAD was enriched for \textit{Antigen processing and presentation} (Fig.~\ref{fig:validation_combined}f), reflecting the distinct immune landscape and immunogenicity often observed in adenocarcinoma subtypes. The significant separation in activity scores demonstrates that LaCoGSEA generalizes well across tissues, capturing both broad proliferative signals and nuanced, tissue-specific metabolic and microenvironmental adaptations.

\subsection{Cross-dataset consistency of pathway ranking} 

To assess the model's ability to prioritize experimentally validated mechanisms, we benchmarked LaCoGSEA against seven baseline methods across five independent datasets, including cancer (DLBCL) and non-cancer pathologies (Alzheimer’s, Trauma, Heart failure, Liver rejection) (Fig.~\ref{fig:cross_performance_five_datasets}). For each dataset, we measured the rank of the the literature-supported disease-associated pathway among all KEGG pathways (FDR $< 0.05$), as well as the coverage, defined as the percentage of datasets in which the target pathway was successfully retrieved within the top $N$ ranks. LaCoGSEA consistently achieved top performance, with a mean target rank of 17.5 and a coverage of 95\% across all datasets (Fig.~\ref{fig:cross_performance_five_datasets}c), substantially outperforming Standard DE (mean rank = 73.6, coverage = 29\%) and linear PCA-based approaches.  

We further compared LaCoGSEA with gradient-based XAI methods, such as SHAP and DeepLIFT, which yielded mean ranks between 44.7 and 77.0, with substantially lower coverage. While gradient-based methods are effective for feature attribution, they are designed to identify a minimal subset of features sufficient for predicting the output. Consequently, they tend to assign high importance to a few dominant driver genes. In contrast, by leveraging the global correlation structure between genes and latent dimensions, LaCoGSEA retrieves the full spectrum of co-regulated genes, reflecting pathway-level biological redundancy. Additionally, the correlation-based metric is computationally efficient compared to gradient-based attribution methods, which require backpropagation or sampling procedures.

Finally, we evaluated model robustness on limited data (N=30; Fig.~\ref{fig:cross_performance_five_datasets}a). Despite the limited sample size, LaCoGSEA demonstrated superior sensitivity compared to all baseline methods. The model assigned the highest priority (Rank 1) to the \textit{Notch signaling pathway} and \textit{Glycolysis Gluconeogenesis}, significantly outperforming linear PCA baselines which failed to prioritize these critical targets effectively. Most notably, LaCoGSEA was the only method capable of identifying the \textit{MAPK signaling pathway} and the \textit{Adipocytokine signaling pathway}. While these subtle, tissue-specific signaling cascades were completely missed by Standard DE, PCA, and XAI approaches, LaCoGSEA successfully retrieved them, underscoring the framework's unique capability to resolve complex, low-signal pathway structures in small-cohort clinical studies.

\section{Discussion and Conclusion}

Standard approaches for pathway enrichment analysis largely fall into two categories: supervised methods that rely on differential expression rankings derived from known labels, and unsupervised single-sample methods that compute statistics based on intra-sample gene ranks. While effective, the former requires prior biological knowledge, and the latter may be sensitive to technical noise and dropouts inherent in transcriptomic data. In this study, we introduce LaCoGSEA, a novel framework that bridges this methodological gap by establishing an ``unsupervised pre-ranking" mechanism. Unlike existing approaches, LaCoGSEA utilizes a deep autoencoder to capture non-linear data manifolds and leverages the global correlation between genes and latent dimensions to generate robust pre-ranked gene lists. Specifically, while the autoencoder employs non-linear transformations to unfold complex genomic manifolds, the resulting latent space effectively disentangles these high-dimensional dependencies. Consequently, Pearson correlation becomes mathematically appropriate to quantify the alignment between genes and these unravelled latent features. This strategy allows us to deploy the statistically rigorous GSEA algorithm in a fully unsupervised setting, transforming latent dimensions into biologically interpretable pathway signatures without the need for manual curation or clinical labels. 

A notable finding of our study is the superior performance of the simple Pearson correlation metric over gradient-based explainable AI (XAI) methods, such as SHAP and DeepLIFT, across benchmark datasets. This observation highlights a fundamental misalignment between the mathematical objectives of current XAI techniques and the biological nature of pathway regulation. Gradient-based attribution methods are typically optimized to identify a minimal subset of features sufficient for prediction, thereby favoring sparse explanations. However, biological systems are inherently redundant and densely coordinated, with pathway activity arising from the concerted expression of large gene networks rather than isolated driver genes. As a consequence, sparsity-enforcing XAI methods may underrepresent co-regulated genes that collectively define pathway-level signals. In contrast, the global correlation metric employed by LaCoGSEA captures the full co-regulation structure of gene sets, aligning more closely with the conceptual foundation of enrichment analysis. Collectively, these results challenge the implicit assumption that sparsity necessarily equates to interpretability in biological pathway analysis and underscore the importance of program-level, rather than feature-level, explanations in unsupervised transcriptomic modeling.

The saturation analysis highlights a key advantage of the deep learning over linear methods: the ability to learn non-linear representations without strict orthogonality constraints. Linear methods enforce statistical independence, which can fragment coherent biological signals arising from overlapping pathways. By relaxing this constraint, LaCoGSEA preserves these complex dependencies, resolving subtle phenotypes that remain obscured in linear projections. Furthermore, the framework effectively maps abstract dimensions to tangible biological processes, such as metabolic trade-offs, without requiring massive training datasets. Its high prioritization accuracy in small-cohort studies indicates that the model learns data-efficient representations, supporting potential clinical applications where data is scarce.

Despite its strengths, LaCoGSEA has several limitations and practical considerations. Although we employ a standard autoencoder for stability, the correlation-based pre-ranking mechanism itself is model-agnostic and extendable to other non-linear representation models. First, as an unsupervised method, the discovered pathways represent the dominant sources of variation in the dataset, which may not always correspond to clinical outcomes of interest without downstream association modeling. Second, although the framework maps latent dimensions to known pathways, interpretation can be challenging for highly overlapping or pleiotropic gene sets, and the resolution of novel or tissue-specific mechanisms is limited by the completeness of reference databases (e.g., KEGG, GO). Finally, a practical consideration concerns the choice of the latent dimensionality ($D$) in LaCoGSEA. As with most representation learning frameworks, the appropriate dimensionality depends on dataset size, signal complexity, and the biological resolution of interest. In practice, we observe that lower-dimensional representations capture dominant programs in small cohorts, whereas higher-dimensional embeddings enable finer-grained decomposition in larger datasets. This trade-off reflects a general balance between model capacity and interpretability rather than a fixed optimal choice.
 
In conclusion, LaCoGSEA presents a new framework for unsupervised pathway analysis. By combining non-linear feature learning of deep autoencoders with a statistically robust, correlation-based pre-ranking mechanism, it overcomes limitations of both linear decomposition and single-sample scoring. Our framework offers a highly interpretable tool for dissecting complex disease heterogeneity, enabling researchers to uncover meaningful biological mechanisms from high-dimensional transcriptomic data with improved clarity and reproducibility.


\section{Author contributions}
Zhiwei Zheng (Conceptualization, Methodology, Software, Validation, Formal analysis, Investigation, Data curation, Visualization, Writing—original draft) and Kevin Bryson (Conceptualization, Methodology, Resources, Writing—review \& editing, Supervision, Project administration).

Conflict of interest: None declared.

\section{Funding}
None.

\section{Data availability}
The RNA-seq datasets analyzed in this study are publicly accessible: the SCAN-B dataset is available via Mendeley Data (\url{https://doi.org/10.17632/yzxtxn4nmd.4}); and the METABRIC dataset can be accessed through the European Genome-phenome Archive (EGA) \url{https://ega-archive.org/studies/EGAS00000000083}). The TCGA-Lung dataset was obtained from the UCSC Xena Browser (\url{https://xenabrowser.net/datapages/?cohort=TCGA%20Pan-Cancer%20(PANCAN)}). For five GEO datasets, they could be accessed in (\url{https://www.ncbi.nlm.nih.gov/geo/}) by search their GEO number (GSE10846, GSE48350, GSE11375, GSE126848, GSE116250).

\bibliographystyle{apalike}
\bibliography{reference}


\end{document}